\documentclass[10pt,twocolumn,letterpaper]{article}

\usepackage{cvpr}
\usepackage{times}
\usepackage{epsfig}
\usepackage{graphicx}
\usepackage{amsmath}
\usepackage{amssymb}
\usepackage{algorithm,algorithmic,alltt}

\usepackage{soul}
\usepackage{multirow}
\usepackage{subfig}

\usepackage[pagebackref=true,breaklinks=true,letterpaper=true,colorlinks,bookmarks=false]{hyperref}

\cvprfinalcopy 

\def\cvprPaperID{1249} 

\ifcvprfinal\fi
\begin{document}

\hyphenpenalty=500

\title{CNN in MRF: Video Object Segmentation via Inference in \\A CNN-Based Higher-Order Spatio-Temporal MRF}


\author{Linchao Bao \quad\quad Baoyuan Wu \quad\quad \ \ Wei Liu\ \ \ \ \ \ \  \\
Tencent AI Lab\\
{\tt\small linchaobao@gmail.com \quad wubaoyuan1987@gmail.com \quad wliu@ee.columbia.edu}
}

\maketitle
\thispagestyle{empty}

\begin{abstract}
This paper addresses the problem of video object segmentation, where the initial object mask is given in the first frame of an input video.
We propose a novel spatio-temporal Markov Random Field (MRF) model defined over pixels to handle this problem.
Unlike conventional MRF models, the spatial dependencies among pixels in our model are encoded by a Convolutional Neural Network (CNN).
Specifically, for a given object, the probability of a labeling to a set of spatially neighboring pixels can be predicted by a CNN trained for this specific object.
As a result, higher-order, richer dependencies among pixels in the set can be implicitly modeled by the CNN.
With temporal dependencies established by optical flow, the resulting MRF model combines both spatial and temporal cues for tackling video object segmentation.
However, performing inference in the MRF model is very difficult due to the very high-order dependencies.
To this end, we propose a novel CNN-embedded algorithm to perform approximate inference in the MRF.
This algorithm proceeds by alternating between a temporal fusion step and a feed-forward CNN step.
When initialized with an appearance-based one-shot segmentation CNN, our model outperforms the winning entries of the DAVIS 2017 Challenge,
without resorting to model ensembling or any dedicated detectors.
\end{abstract}

\section{Introduction}

Video object segmentation refers to a task of extracting pixel-level masks for class-agnostic objects in videos.
This task can be further divided into two settings \cite{perazzi2016benchmark}, namely \emph{unsupervised} and \emph{semi-supervised}.
While the unsupervised task does not provide any manual annotation, the semi-supervised task provides information about objects of interest in the first frame of a video.
In this paper, we focus on the latter task, where the initial masks for objects of interest are provided in the first frame.
The task is important for many applications such as video editing, video summarization, action recognition, \etc.
Note that the semantic class/type of the objects of interest cannot be assumed known and the task is thus class-agnostic.
It is usually treated as a temporal label propagation problem and solved with spatio-temporal graph structures \cite{grundmann2010efficient,perazzi2015fully,TsaiBMVC10,avinash2014seamseg}
like a Markov Random Field (MRF) model \cite{tsai2016video}.
Recent advances on the task show significant improvements over traditional approaches when incorporating deep Convolutional Neural Networks (CNNs) \cite{caelles2017one,perazzi2017learning,voigtlaender2017online,shin2017pixel,cheng2017segflow,jang2017online,jampani2017cvpr}.
Despite the remarkable progress achieved with CNNs, video object segmentation is still challenging
when applied in real-world environments. One example is that even the top performers \cite{caelles2017one,perazzi2017learning}
on the DAVIS 2016 benchmark \cite{perazzi2016benchmark} show significantly worse performance on the more challenging DAVIS 2017 benchmark \cite{PontTuset2017davischallenge}, where interactions between objects, occlusions, motions, object deformation, \etc, are more complex and frequent in the videos.

Reviewing the top performing CNN-based methods and traditional spatio-temporal graph-based methods, there is a clear gap between the two lines.
The CNN-based methods usually treat each video frame individually or only use simple heuristics to propagate information along the temporal axis,
while the well established graph-based models cannot utilize the powerful representation capabilities of neural networks. In order to fully exploit
the appearance/shape information about the given objects, as well as the temporal information flow along the time axis, a better solution should be
able to combine the best from both. For example, built on the top-performing CNN-based methods \cite{caelles2017one,perazzi2017learning}, there should be
a temporal averaging between the CNN outputs of an individual frame and its neighboring frames, so that the segmentation results are temporally consistent.
The temporal averaging, however, is heuristic and likely to degrade the segmentation performance due to outliers. A more principled method should be
developed. In this paper, we propose a novel approach along this direction.

Specifically, we build a spatio-temporal MRF model over a video sequence, where each random variable represents the label of a pixel.
While the pairwise temporal dependencies between random variables are established using optical flow between neighboring frames,
the spatial dependencies in our model are not modeled as pairwise potentials like conventional MRF models \cite{shotton2009textonboost,koller2009probabilistic,baoyuanwu-cvpr-2013,baoyuanwu-iccv-2013,baoyuanwu-pr-2017}.
The problem of spatial pairwise potential is that it has very restricted expressive power and thus cannot model complicated
dependencies among pixels in natural images. Some higher-order potentials are proposed to incorporate learned patterns \cite{roth2005fields} or enforce
label consistency in pre-segmented regions \cite{kohli2009robust,arnab2016higher}. Yet the expressive power of them is still rather limited.

In our model, we instead use a CNN to encode even higher-order spatial potentials over pixels.
Given a labeled object mask in the first frame, we can train a mask refinement CNN for the object
to refine a coarse mask in a future frame.
Assuming that the mask refinement CNN is so reliable that it can consistently refine a coarse
mask to a better one and keep a good mask unchanged,
we can define an objective function based on the CNN to assess a given mask as a whole.
Then the spatial potential over the pixels within a frame can be defined using the CNN-based function.
In this case, more complicated dependencies among pixels can be represented for the object.
As a result, the MRF model will enforce the inference result
in each frame to be more like the specific object.
Yet, the inference in the resulting MRF model is very difficult due to the CNN-based potential function.
In this paper, we overcome the difficulty by proposing a novel approximate inference algorithm for the MRF model.
We first decouple the inference problem into two subproblems by introducing an auxiliary variable.
Then we show that one subproblem involving the CNN-based potential function can be approximated by a feed-forward pass of the mask refinement CNN.
Consequently, we do not even need to explicitly compute the CNN-based potential function during the inference.
The entire inference algorithm alternates between a temporal fusion step and a feed-forward pass of the CNN.
When initialized with a simple one-shot segmentation CNN \cite{caelles2017one},
our algorithm shows outstanding performance on challenging benchmarks like the DAVIS 2017 \emph{test-dev} dataset \cite{PontTuset2017davischallenge}.

\subsection{Related Work}

\noindent\textbf{Video Object Segmentation} We briefly review recent work focusing on the semi-supervised setting.
The task is usually formulated as a temporal label propagation problem.
Spatio-temporal graph-based methods tackle the problem by building up graph structures over
pixels \cite{TsaiBMVC10}, patches \cite{avinash2014seamseg}, superpixels \cite{grundmann2010efficient,tsai2016video}, 
or even object proposals \cite{perazzi2015fully} to infer the labels for subsequent frames.
The temporal connections are established using regular spatio-temporal lattices \cite{marki2016bilateral}, optical flow \cite{grundmann2010efficient},
or other similar techniques like nearest neighbor fields \cite{avinash2014seamseg}.
Some methods even build up long-range connections using appearance-based methods \cite{perazzi2015fully}.
Among these methods, some algorithms infer the labels using greedy strategies by only considering two or more neighboring frames one time \cite{fan2015jumpcut,marki2016bilateral,tsai2016video},
while other algorithms strive to find globally optimal solutions by considering all the frames together \cite{TsaiBMVC10,grundmann2010efficient,perazzi2015fully}.

Although various nicely designed models and algorithms are proposed to tackle the problem,
deep learning shows overwhelming power when introduced to this area. It is shown that a merely appearance-based CNN named OSVOS \cite{caelles2017one}, trained on the first frame
of a sequence and tested on each subsequent frame individually, achieves significant improvements over top-performing traditional methods ($79.8\%$ vs $68.0\%$ accuracy on the DAVIS 2016 dataset \cite{perazzi2016benchmark}). 
A concurrent work named MaskTrack \cite{perazzi2017learning} achieves similar performance by employing a slightly different CNN where the mask of
a previous frame is fed to the CNN as an additional channel besides the RGB input image.
Some other CNN-based methods also demonstrate pretty nice results \cite{jang2017online,jampani2017cvpr,shin2017pixel,cheng2017segflow,voigtlaender2017online}.
Among these methods, an online adaptation version of OSVOS, namely OnAVOS \cite{voigtlaender2017online}, achieves the best performance ($86.1\%$ accuracy) on the DAVIS 2016 dataset.

Since the best performance on the DAVIS 2016 dataset tends to be saturated, the authors of the dataset released a larger, more challenging dataset,
namely DAVIS 2017 \cite{PontTuset2017davischallenge}, to further push the research in video object segmentation for more practical use cases.
The new dataset adds more distractors, smaller objects and finer structures, more occlusions and faster motions, \etc.
Hence the top performer on DAVIS 2016 performs much worse when it comes to the new dataset. For example, the accuracy of OnAVOS \cite{voigtlaender2017online}
on DAVIS 2016 is $86.1\%$, while its accuracy drops to $50.1\%$ on the DAVIS 2017 \emph{test-dev} dataset.
Although the best performance (on the \emph{test-dev} dataset) is further improved to around $66\%$ during the DAVIS 2017 Challenge
by Li \etal \cite{DAVIS2017-1st} and the LucidTracker \cite{DAVIS2017-2nd},
the score is achieved with engineering techniques like model ensembling, multi-scale training/testing, dedicated object detectors,
\etc.
We show in this paper that our proposed approach can achieve better performance without resorting to these techniques.



\noindent\textbf{CNN + MRF/CRF} The idea of combining the best from both CNN and MRF/CRF is not new.
We here briefly review some attempts to combine CNN and MRF/CRF for the segmentation task. For a more thorough review please refer to \cite{arnabconditional2018review}.
The first idea to take advantage of the representation capability of CNN and the fine-grained probabilistic modeling capability of MRF/CRF
is to append an MRF/CRF inference to a CNN as a separate step. For example, the semantic segmentation framework DeepLab \cite{chen2017deeplabpami}
utilizes fully-connected CRFs \cite{krahenbuhl2011efficient} as a post-processing step to improve the semantic labelling results produced by a CNN,
similar to performing an additional edge-preserving filtering \cite{gastal2011domain,bao2014tree,bao2012icprdehaze,bao2014robust} on the segmentation masks.
The video object segmentation method by Jang and Kim \cite{jang2017online} performs MRF optimization to fuse the outputs of a triple-branch CNN.
However, the loosely-coupled combination cannot fully exploit the strength of MRF/CRF models. Schwing and Urtasun \cite{schwing2015fully}
proposed to jointly train CNN and MRF by back-propagating gradient obtained during the MRF inference to CNN. Unfortunately, the approach
does not show clear improvements over the separated training scheme.
Arnab \etal \cite{arnab2016higher} successfully demonstrated performance gains via a joint training of CNN and MRF, even with higher-order
potentials modeled by object detection or superpixels. Note that their focus is on the back-propagation of high-order potentials during the joint training,
while our work focuses on the higher-order modeling with CNNs.
The CRF-RNN work \cite{zheng2015conditional} formulates the mean-field approximate inference for CRFs as a Recurrent Neural Network (RNN)
and integrates it with a CNN to obtain an end-to-end trainable deep network, which shows an outstanding performance boost in an elegant way.
Taking one step further, the Deep Parsing Network (DPN) \cite{liu2017deep} is designed to approximate the mean-field inference for MRFs in one pass.
The above work demonstrates promising directions of using neural networks to approximate the inference of MRFs,
which is different from our work that is trying to model higher-order potentials in MRFs with CNNs.


\subsection{Contributions}

The main contributions of this paper are as follows:

\begin{itemize}
  \item We propose a novel spatio-temporal Markov Random Field (MRF) model for the video object segmentation problem.
  The novelty of the model is that the spatial potentials are encoded by CNNs trained for objects of interest,
  so higher-order dependencies among pixels can be modeled to enforce the holistic segmentation of object instances.

  \item We propose an effective iterative algorithm for video object segmentation.
  The algorithm alternates between a temporal fusion operation and a feed-forward CNN to progressively refine the segmentation results.
  Initialized with an appearance-based one-shot video object segmentation CNN, our algorithm achieves state-of-the-art performance on public benchmarks.
\end{itemize}

\section{Model}\label{sec.ourmodel}

We start by considering the case of the single object segmentation, where the goal is to label pixels as binary values.
Handling multiple objects is described in Sec. \ref{sec.implementationdetails}.
Note that in the semi-supervised setting, the ground-truth object mask for the first frame of a video is given.

\subsection{Notations \& Preliminaries}

We define a discrete random field $\mathbf{X}$ over all the pixels $\mathcal{V} = \{1,2,...,N\}$ in a video sequence.
Each random variable $X_i \in \mathbf{X}$ is associated with a pixel $i \in \mathcal{V}$ and takes a value $x_i$ from the label set $\mathcal{L}=\{0,1\}$.
We use $\mathbf{x}$ to denote a possible assignment of labels (namely a \emph{labeling} or a \emph{configuration}) to the random variables in $\mathbf{X}$.
The data of video frames is denoted as $\mathbf{D}$.
Denoting a clique in the field by $c$ and the set of variables in that clique by $\mathbf{x}_c$,
the distribution of the random variables in the field can be written as a product of potential functions
over the maximal cliques \cite{bishop2006pattern}
\begin{equation}\label{eq.mrffactors}
  p(\mathbf{x}|\mathbf{D}) = \frac{1}{Z} \prod_c \psi_c(\mathbf{x}_c|\mathbf{D}) = \frac{1}{Z} \prod_c \mathrm{exp} \{ - E_c(\mathbf{x}_c|\mathbf{D}) \},
\end{equation}
where $Z$ is the normalization constant and $E_c(\mathbf{x}_c|\mathbf{D})$ is the energy function corresponding to the potential function $\psi_c(\mathbf{x}_c|\mathbf{D})$.
Our goal is to infer the maximum a posteriori (MAP) labeling $\mathbf{x}^*$ of the random field as
\begin{equation}\label{eq.mapmrf}
  \mathbf{x}^* = \arg\max_\mathbf{x} \log p(\mathbf{x}|\mathbf{D}) = \arg\min_\mathbf{x} \sum_c  E_c(\mathbf{x}_c|\mathbf{D}).
\end{equation}
By defining the graph structures of the random field and their associated energy functions,
the MAP labeling can be obtained via minimizing the total energy in the field.
Note that the energy functions defined in our model will be conditioned on the data $\mathbf{D}$.
To be concise, we will drop $\mathbf{D}$ in the notations hereafter.


\subsection{Model Structures \& Energies}

The total energy in our model is defined as follows
\begin{equation}\label{eq.totalenergy}
  E(\mathbf{x}) = \sum_{i \in \mathcal{V}} E_u(x_i) + \sum_{(i,j) \in \mathcal{N}_T} E_t(x_i, x_j) + \sum_{c \in \mathcal{S}} E_s(\mathbf{x}_c),
\end{equation}
where $E_u$ is the unary energy, and $E_t$ and $E_s$ are the energies associated with temporal and spatial dependencies, respectively.
The notation $\mathcal{N}_T$ refers to the set of all temporal connections,
while $\mathcal{S}$ refers to the set of all spatial cliques.
The concrete definitions are as follows.

The unary energy is defined by the negative log likelihood of the labeling for each individual random variable as
\begin{equation}\label{eq.unaryenergy}
  E_u(x_i) = - \theta_u \log p(X_i=x_i),
\end{equation}
where $\theta_u$ is to balance the weight of this term with other energy terms.

\begin{figure}[t!]
    \vspace{-1mm}
    \includegraphics[width=0.48\textwidth]{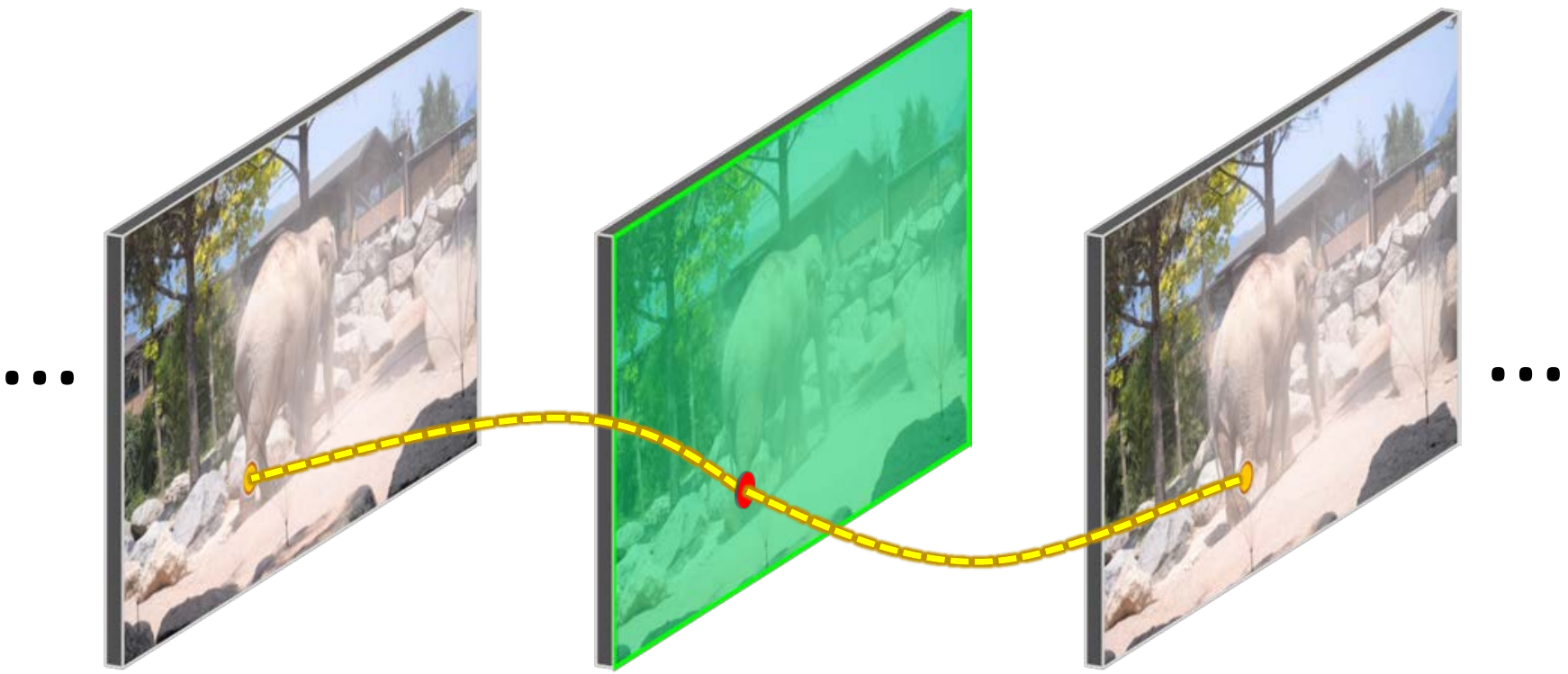}
    \caption{The spatio-temporal dependencies for a pixel (in red) in our model.
    The temporal dependencies are established by optical flow (indicated by yellow dashed lines).
    The spatial dependencies are modeled by a CNN, as shown in the center frame where the green shaded region
    indicates pixels belonging to the same spatial clique as the red pixel (in this case it indicates all the pixels within the same image as the red pixel).
    Best viewed in color.}
    \vspace{-2mm}
    \label{fig.cnnmrffig}
\end{figure}

The set of temporal connections $\mathcal{N}_T$ is established using semi-dense optical flow, such that each pixel is only connected
to pixels in neighboring frames when the motion estimation is reliable enough.
We use a forward-backward consistency check to filter reliable motion vectors \cite{bao2014cvpreppm,bao2014fast,jiang2012trajectory,jiang2015human}.
The yellow dashed lines in Fig. \ref{fig.cnnmrffig} show an example of an one-step temporal dependencies for the red pixel.
Note that the one-step temporal dependencies can be further extended to $k$-step temporal dependencies by directly computing
optical flow between a frame and the frame that is $k$ frames away. We use $k \leqslant 2$ $(k \neq 0)$ in our model,
which means for a certain frame $t$, all the following links are established:
$\{t \longleftrightarrow t-2\}$, $\{t \longleftrightarrow t-1\}$, $\{t \longleftrightarrow t+1\}$, $\{t \longleftrightarrow t+2\}$.
As a result, each pixel is connected to at most $4$ temporal neighbors (note that some invalid connections are removed by the forward-backward consistency check).
The temporal energy function is defined as
\begin{equation}\label{eq.temporalenergy}
  E_t(x_i, x_j) = \theta_t w_{ij} (x_i - x_j)^2,
\end{equation}
where $\theta_t$ is a balancing parameter for this term
and $w_{ij}$ is a data-dependent weight to measure the confidence of the temporal connection between variables $X_i$ and $X_j$.
The energy encourages a temporally consistent labeling when the temporal connection is confident.


For spatial dependencies, we define all the pixels in a frame as a clique,
in which the labeling for each pixel depends on all other pixels in the same frame (shown as the green shaded region in Fig. \ref{fig.cnnmrffig}).
In order to construct a spatial energy function defined over all the pixels in a frame,
we need to design an energy function $f(\cdot)$ that can assess the quality of a given mask $\mathbf{x}_c$ as a whole.
Ideally, it is easy to construct the function $f(\cdot)$ if the ground-truth mask $\mathbf{x}_c^*$ of an input mask $\mathbf{x}_c$ is given.
For example, we can define $f(\cdot)$ 
\begin{equation}\label{eq.spatialenergyfuncgroundtruth}
  f(\mathbf{x}_c) = \| \mathbf{x}_c - \mathbf{x}_c^* \|_2^2,
\end{equation}
which gives lower energies to masks that are more similar to the ground-truth mask.
However, $\mathbf{x}_c^*$ is unknown and indeed what we need to solve for.
We here resort to a feed-forward CNN to approximate $\mathbf{x}_c^*$ and define $f(\cdot)$ as follows
\begin{equation}\label{eq.spatialenergyfuncgcnn}
  f(\mathbf{x}_c) = \| \mathbf{x}_c - g_{\mathrm{CNN}}(\mathbf{x}_c) \|_2^2,
\end{equation}
where $g_{\mathrm{CNN}}(\cdot)$ is a mask refinement CNN that accepts as input a given mask $\mathbf{x}_c$ and outputs a refined mask.
Note that the operator $g_{\mathrm{CNN}}(\cdot)$ here is a feed-forward pass of a CNN.
Intuitively, the above definition assigns a lower energy to a mask whose mapping through $g_{\mathrm{CNN}}(\cdot)$ is more similar to itself.
With a well-trained $g_{\mathrm{CNN}}(\cdot)$ that can reliably refine a coarse mask to a better one and keep a good mask unchanged,
the function $f(\cdot)$ could assign better masks lower energies.
Fortunately, it is shown that such a CNN can be trained in a two-stage manner using the first frame of a given video
and performs very reliably during the inference for the following frames \cite{perazzi2017learning}.
We leave the detailed discussion of $g_{\mathrm{CNN}}(\cdot)$ in the next section.
We here define the spatial energy in Eq. \eqref{eq.totalenergy} as
\begin{equation}\label{eq.spatialenergy}
  E_s(\mathbf{x}_c) = \theta_s f(\mathbf{x}_c),
\end{equation}
where $\theta_s$ is a balancing parameter for this term.

The above spatial energy definition has a much more expressive power than traditional pairwise smoothness energies \cite{shotton2009textonboost}, 
higher-order energies enforcing label consistency in pre-segmented regions \cite{kohli2009robust,arnab2016higher},
or energies encouraging labels to follow certain learned patterns \cite{roth2005fields}.
However, the inference in the MRF with the CNN-based energy is very difficult.
In the next section, we present an efficient approximate algorithm for the inference.

\section{Inference}\label{sec.ourinference}

The exact MAP inference in MRF models is NP-hard in general \cite{koller2009probabilistic}.
The higher-order energy function in our model makes the inference problem intractable.
Even with efficient approximate algorithms like belief propagation or mean-field approximation,
finding a solution minimizing Eq. \eqref{eq.totalenergy} is still computationally infeasible,
due to the very high-order spatial cliques.
Intuitively, each time evaluating the total energy in the MRF, a feed-forward CNN pass for every frame in the video is required.
The computational cost quickly becomes unaffordable as the number of energy evaluations grows.

In order to make the problem tractable, we decouple the temporal energy $E_t$ and spatial energy $E_s$
by introducing an auxiliary variable $\mathbf{y}$,
and minimize the following approximation of Eq. \eqref{eq.totalenergy} instead:
\begin{align}\label{eq.decoupleauxiliary}
  \hat{E}(\mathbf{x}, \mathbf{y}) &= \sum_{i \in \mathcal{V}} E_u(x_i) + \sum_{(i,j) \in \mathcal{N}_T} E_t(x_i, x_j) \nonumber\\
  & + \frac{\beta}{2} \|\mathbf{x} - \mathbf{y}\|_2^2  + \sum_{c \in \mathcal{S}} E_s(\mathbf{y}_c),
\end{align}
where $\beta$ is a penalty parameter such that $\mathbf{y}$ is a close approximation of $\mathbf{x}$.
Eq. \eqref{eq.decoupleauxiliary} can be minimized by alternating steps updating either $\mathbf{x}$ or $\mathbf{y}$ iteratively.
Specifically, in the $k$-th iteration, the two updating steps are:
\begin{enumerate}
  \setlength{\itemsep}{0pt}
  \setlength{\parskip}{0pt}
  \item with $\mathbf{y}$ fixed, update $\mathbf{x}$ by
  \begin{align}\label{eq.decoupleauxiliarystep1}
  \mathbf{x}^{(k)} \leftarrow \arg\min_{\mathbf{x}} \hat{E}(\mathbf{x}, \mathbf{y}^{(k-1)}),
  \end{align}
  \item with $\mathbf{x}$ fixed, update $\mathbf{y}$ by
  \begin{align}\label{eq.decoupleauxiliarystep2}
  \mathbf{y}^{(k)} \leftarrow \arg\min_{\mathbf{y}} \hat{E}(\mathbf{x}^{(k)}, \mathbf{y}).
  \end{align}
\end{enumerate}
Note that Eq. \eqref{eq.decoupleauxiliarystep1} in step 1 is essentially the regularized total energy
in Eq. \eqref{eq.totalenergy} ignoring spatial dependencies,
while Eq. \eqref{eq.decoupleauxiliarystep2} in step 2 only considers spatial dependencies for each frame $c$.
The two steps are essentially performing \emph{temporal fusion} and \emph{mask refinement}, respectively.




Solving step 1 requires solving a large quadratic integer programming problem with an $N \times N$ Laplacian matrix,
where $N$ is the total number of pixels in a video.
For efficiency considerations, we here resort to a classical iterative method named
Iterated Conditional Modes (ICM) \cite{bishop2006pattern} to find an approximate solution of step 1.
Specifically, each time one random variable $X_i$ is updated to minimize Eq. \eqref{eq.decoupleauxiliarystep1},
fixing the rest of the random variables $X$.
The original ICM algorithm repeats the variable updating until converged.
In our algorithm, we only perform the updating for a fixed number $L$ of iterations, as shown in Algorithm \ref{alg.iterativealg}.

The updating in step 2 can be performed for each frame $c$ individually, that is
\begin{align}\label{eq.step2opt}
    \mathbf{y}^{(k)}_c \leftarrow \arg\min_{\mathbf{y}_c} \Big\{ \frac{\beta}{2} \|\mathbf{x}_c^{(k)} - \mathbf{y}_c\|_2^2  + E_s(\mathbf{y}_c) \Big\}.
\end{align}
Note that the problem is highly non-convex due to the CNN-based energy function $E_s$.
Intuitively, step 2 is to refine a given mask $\mathbf{x}_c^{(k)}$ such that the output mask $\mathbf{y}_c^{(k)}$ is better
in terms of $E_s$ and at the same time not deviates too much from the input.
Directly solving the optimization problem in Eq. \eqref{eq.step2opt} is difficult,
we instead approximate this step by simply using $g_{\mathrm{CNN}}(\cdot)$ to update $\mathbf{y}_c$:
\begin{align}
    \mathbf{y}^{(k)}_c \leftarrow g_{\mathrm{CNN}}(\mathbf{x}_c^{(k)}),
\end{align}
which we find in the experiments can make the objective function in Eq. \eqref{eq.step2opt} non-increasing in most cases
($99\%$ of more than 3000 frames in the DAVIS 2017 validation set when $\theta_s = \beta$).
As a result, the overall algorithm alternating between the above two steps, as described in Algorithm \ref{alg.iterativealg},
ensures the non-increasing of the total energy in Eq. \eqref{eq.decoupleauxiliary} in each iteration.
We show experimentally in Sec. \ref{sec.ourexperiments} that the algorithm converges after a few iterations.

\begin{algorithm}[t!]
\caption{Our Inference Algorithm}
\begin{algorithmic}
\STATE \textbf{Parameters}: number of outer iterations $K$, number of inner iterations $L$, number of pixels $N$, and number of frames $C$.
\STATE \textbf{Initialization}: initial labeling $\mathbf{x}^{(0)}=\mathbf{y}^{(0)}$.
\FOR{$k$ from $1$ to $K$}
    \STATE \emph{-- Temporal Fusion Step (TF) --}
    \STATE $\mathbf{x}^{(k,0)} \leftarrow \mathbf{x}^{(k-1)}$
    \FOR{$l$ from $1$ to $L$}
        \FOR{$i$ from $1$ to $N$}
            \STATE $x_i^{(k,l)} \leftarrow  \arg\min_{x_i} \Big\{ \frac{\beta}{2} (x_i - y_i^{(k-1)})^2 + E_u(x_i)$
            \STATE \quad\quad\quad\quad\quad\quad $+  \sum_{(i,j) \in \mathcal{N}_T} E_t(x_i, x_j^{(k,l-1)}) \Big\}$
        \ENDFOR
    \ENDFOR
    \STATE $\mathbf{x}^{(k)} \leftarrow \mathbf{x}^{(k,L)}$
    \STATE \emph{-- Mask Refinement Step (MR) --}
    \FOR{$c$ from $1$ to $C$}
        \STATE $\mathbf{y}^{(k)}_c \leftarrow g_{\mathrm{CNN}}(\mathbf{x}_c^{(k)})$
    \ENDFOR
\ENDFOR
\STATE \textbf{Output}: Binarize $\mathbf{y}^{(K)}$ as the final segmentation masks.
\end{algorithmic}
\label{alg.iterativealg}
\end{algorithm}

Now we discuss the details of the CNN operator $g_{\mathrm{CNN}}(\cdot)$ in our algorithm.
Similar to MaskTrack \cite{perazzi2017learning}, our $g_{\mathrm{CNN}}(\cdot)$ accepts a 4-channel input (RGB image + coarse mask),
and outputs a refined mask.
Also, we train $g_{\mathrm{CNN}}(\cdot)$ in a two-stage manner.
In the first stage, an offline model is trained using object segmentation data available.
Then in the second stage, the offline model is fine-tuned using the ground-truth mask in the first frame of a given video.
During the training, the input mask to the CNN is a contaminated version of the ground-truth mask
with data augmentation techniques like non-rigid deformation.
Note that the two-stage training is performed before our inference algorithm.
During inference, the operator $g_{\mathrm{CNN}}(\cdot)$ in Algorithm \ref{alg.iterativealg} is only a feed-forward pass of the CNN.

The CNN trained in this way can partially encode the appearances of an object of interest.
It endows our algorithm with the ability to recover missing parts of an object mask when occlusions happen.
Even in the case that there are re-appearing objects after completely occluded, our algorithm can recover high-quality
object masks given a poor likelihood obtained from an appearance-based method like OSVOS \cite{caelles2017one}.
In fact, we will show in the next section that our algorithm achieves outstanding performance on challenging datasets
where heavy occlusions are common.

\section{Experiments}\label{sec.ourexperiments}

\begin{figure*}[!t]
    \vspace{-3mm}
    \centerline{\subfloat[Baseline]{\label{fig.ablationexample.baseline}
        \includegraphics[width=0.19\textwidth]{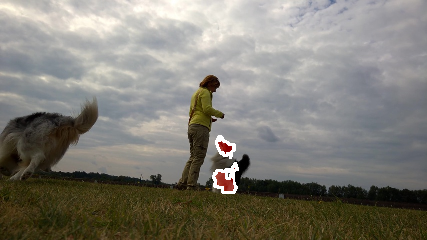}}
        \subfloat[Baseline+TF]{\label{fig.ablationexample.tf}
        \includegraphics[width=0.19\textwidth]{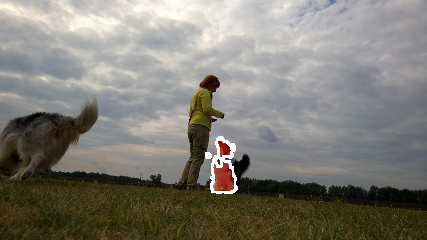}}
        \subfloat[Baseline+MR]{\label{fig.ablationexample.mr}
        \includegraphics[width=0.19\textwidth]{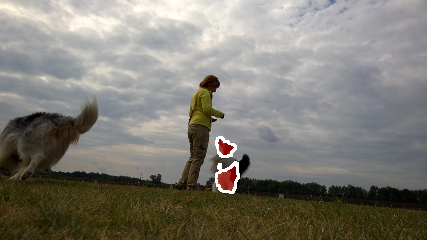}}
        \subfloat[Baseline+TF\&MR]{\label{fig.ablationexample.tfmr}
        \includegraphics[width=0.19\textwidth]{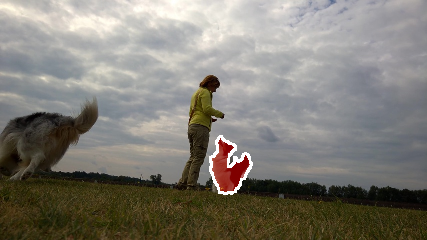}}
        \subfloat[Ground-Truth]{\label{fig.ablationexample.gt}
        \includegraphics[width=0.19\textwidth]{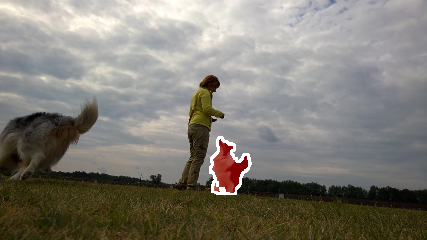}}}\vspace{-8pt}
    \caption{An example of the ablation study experiments.
    Performing TF or MR individually can only yield limited improvements over the baseline method, as shown in (b) and (c).
    With both TF and MR enabled, the quality of the segmentation result gets largely improved, as shown in (d).}
    \vspace{-4mm}
    \label{fig.ablationexample}
\end{figure*}

\subsection{Implementation Details}\label{sec.implementationdetails}

\noindent\textbf{Initialization \& Pixel Likelihood} We use OSVOS \cite{caelles2017one}
to obtain the initial labeling and pixel likelihood for all frames.
As OSVOS tends to produce false-positive results, especially when there are multiple similar objects,
we weight the response map output from OSVOS with a Gaussian centered at the most likely location of the target object
predicted by a simple linear motion model \cite{breitenstein2009robust,luo2014multiple}.
The weighted response map in each frame is then combined (using \texttt{max} at each pixel) with the response map warped from the preceding frame.
Then the fused response map is binarized as the initial labeling of our algorithm.
To obtain the pixel likelihood $p(X_i=x_i)$ in Eq. \eqref{eq.unaryenergy},
we use the initial foreground region imposed with a dilated uncertain region similar to \cite{voigtlaender2017online},
where we assign the likelihood for pixels in the foreground region as $p(X_i=1)=0.99$
and the likelihood for pixels in the uncertain region as a Gaussian peaked with probability $p(X_i=1)=0.7$.


\noindent\textbf{CNN Implementation} We use the Caffe-based DeepLab framework \cite{chen2017deeplabpami} to implement our mask refinement CNN $g_{\mathrm{CNN}}(\cdot)$.
The backbone net is a VGG-Net \cite{Simonyan15vgg} with the input data layer modified to 4-channel (RGB image + 1-channel binary mask).
We add additional skip connections from intermediate pooling layers to a final output convolutional layer to enable multi-level feature fusion.
The input image to our CNN is cropped around the object using the labeling from a previous iteration
and then resized to $513 \times 513$. In our experiments, we train the offline model using the DAVIS 2017 training set \cite{PontTuset2017davischallenge}
for 50K iterations with a batch size of $10$ and a learning rate of $10^{-4}$ (with ``poly'' policy),
where the initial model weights are obtained from DeepLabv2\_VGG16 pre-trained on PASCAL VOC.
The training data consists of 60 video clips with all frames annotated in pixel-level.
We use the optical flow warped mask of a previous frame as contaminated input for each frame during offline training.
For a given test video, the offline model is fine-tuned for 2K iterations using the ground-truth mask of the first frame,
augmented using a simplified version of Lucid data dreaming \cite{DAVIS2017-2nd}.
We provide the model definition file in supplementary material.

\noindent\textbf{Handling Multiple Objects} When there are multiple objects to be segmented in a video,
we handle each object individually in each iteration and deal with overlapped regions before starting the next iteration.
Overlapped regions are divided into connected pixel blobs and each blob is assigned to a label that minimizes Eq. \eqref{eq.decoupleauxiliarystep1} for the blob.

\noindent\textbf{Other Settings} We use FlowNet2 \cite{ilg2017flownet} to compute optical flow in our implementation.
In the case that \texttt{NaN} error happens, we instead use the TV-L1 Split-Bregman optical flow GPU implementation provided by Bao \etal \cite{bao2014comparison}.
The temporal confidence weighting $w_{ij}$ in Eq. \eqref{eq.temporalenergy} is obtained
by incorporating a decaying frame confidence as the frame index grows, which is $w_{ij}=\xi_{c_i} \xi_{c_j}$,
where $\xi_c = \max (0.9^{c-1}, 0.3)$ for frame $c$ ranging from $1$ to number of frames in a video.
The energy balancing parameters in Eq. \eqref{eq.totalenergy} are set to $\theta_u=\theta_t=1$.
The decoupling penalty parameter in Eq. \eqref{eq.decoupleauxiliary} is initially set to $\beta = 1.5$, multiplied by $1.2$ in each iteration.
The number of inner iterations (\ie, the ICM iterations in temporal fusion) is set to $L=5$. 


\noindent\textbf{Runtime Analysis} The main portion of the runtime is the online Lucid data augmentation and CNN training for a given video.
In our implementation, the data augmentation takes about 1 hour to produce 300 training pairs from the first frame,
and the online training of $g_{\mathrm{CNN}}(\cdot)$ takes about 1 hour for 2K iterations with a batch size of 10 on an NVIDIA Tesla M40 GPU.
The online training of OSVOS takes about 20 minutes for 2K iterations but can be performed in parallel to the training of $g_{\mathrm{CNN}}(\cdot)$.
During inference, the algorithm is actually pretty efficient.
Note that the optical flow for establishing temporal dependencies is only computed once,
with each frame only taking a fraction of a second on GPU \cite{ilg2017flownet,bao2014comparison}.
The temporal fusion step in our algorithm is performed locally and the runtime is almost ignorable.
The mask refinement step is a feed-forward pass of CNN and takes only a fraction of a second on GPU for each frame.

\subsection{Ablation Study}

\begin{table}[t!]\footnotesize
\begin{center}
 \begin{tabular}{ccccccc}
\hline
\multirow{2}{*}{Method} & \multicolumn{2}{c}{Global} & \multicolumn{2}{c}{Region $\mathcal{J}$}    & \multicolumn{2}{c}{Contour $\mathcal{F}$} \\
& Mean & Boost & Mean & Recall & Mean & Recall  \\
\hline
\hline
OSVOS \cite{caelles2017one}  & 0.574 & -- & 0.546 & 0.598 & 0.601 & 0.675\\
Our baseline                 & 0.596 & -- & 0.558 & 0.617 & 0.633 & 0.715\\
+TF$\times 1$                & 0.589 & -0.007  & 0.556 & 0.607 & 0.623 & 0.723\\
+TF$\times 2$                & 0.590 & -0.006  & 0.556 & 0.609 & 0.623 & 0.722\\
+TF$\times 3$                & 0.590 & -0.006  & 0.556 & 0.610 & 0.623 & 0.722\\
+TF$\times 4$                & 0.590 & -0.006  & 0.556 & 0.611 & 0.623 & 0.722\\
+TF$\times 5$                & 0.590 & -0.006  & 0.557 & 0.611 & 0.623 & 0.722\\
+MR$\times 1$                & 0.640 & 0.044  & 0.600 & 0.675 & 0.680 & 0.749\\
+MR$\times 2$                & 0.647 & 0.051  & 0.608 & 0.683 & 0.686 & 0.752\\
+MR$\times 3$                & 0.648 & 0.052  & 0.609 & 0.684 & 0.687 & 0.753\\
+MR$\times 4$                & 0.648 & 0.052  & 0.610 & 0.681 & 0.687 & 0.756\\
+MR$\times 5$                & 0.649 & 0.053  & 0.610 & 0.679 & 0.688 & 0.754\\
+TF\&MR$\times 1$        & 0.692 & 0.096  & 0.652 & 0.728 & 0.732 & 0.822\\
+TF\&MR$\times 2$        & 0.704 & 0.108  & 0.668 & 0.740 & 0.740 & 0.824\\
+TF\&MR$\times 3$        & 0.706 & 0.110  & 0.671 & 0.742 & 0.741 & 0.816\\
+TF\&MR$\times 4$        & 0.707 & 0.111  & 0.672 & 0.744 & 0.742 & 0.820\\
+TF\&MR$\times 5$        & 0.707 & 0.111  & 0.672 & 0.744 & 0.742 & 0.820\\
\hline
\end{tabular}
\end{center}
\vspace{-5mm}
\caption{Ablation study on the DAVIS 2017 validation set. Our baseline is OSVOS enhanced with a linear motion model.
TF represents the temporal fusion step in our algorithm, while MR represents the mask refinement step.
For example, ``TF\&MR$\times 3$'' means that the algorithm is performed for 3 iterations with both TF and MR steps enabled.
The ``Boost'' column shows the performance gain of adding each algorithm variant to the baseline.}
\vspace{-3.5mm}
\label{table.ablationstudy}
\end{table}

We perform ablation study of our algorithm on the DAVIS 2017 validation set \cite{PontTuset2017davischallenge},
which consists of 30 video clips with pixel-level annotations. We use the region similarity in
terms of IoU ($\mathcal{J}$) and contour accuracy ($\mathcal{F}$) to evaluate quality of the results.
Table \ref{table.ablationstudy} shows the evaluation results for our algorithm in different settings.
As a comparison, the results from OSVOS \cite{caelles2017one} (without boundary snapping,
multiple objects conflicts are handled by simply taking the object with the maximum CNN response value at each pixel)
are also listed in the table.
Note that the performance of OSVOS on the DAVIS 2017 dataset is significantly worse than that on the DAVIS 2016 dataset,
since the renewed dataset is much more challenging.
Our baseline implementation is essentially OSVOS (without boundary snapping) enhanced with a linear motion model,
which serves as the initial labeling in our algorithm.

In our experiments, when only performing temporal fusion step (TF), 
we set $\mathbf{y}$ with the updated $\mathbf{x}$ after each iteration in Algorithm \ref{alg.iterativealg}
to propagate variable estimation between iterations. 
Similarly, when only performing mask refinement step (MR), $\mathbf{x}$ is set with the latest $\mathbf{y}$ in each iteration.
From Table \ref{table.ablationstudy} we can see that, performing only TF step will degrade the results of the segmentation.
This is because TF step completely ignores the rich spatial dependencies between variables in a frame, 
and tends to propagate erroneous labeling among neighboring frames. 
On the other hand, performing only MR step can largely boost the baseline method but will
stuck at a performance gain around $5\%$.
With both TF and MR steps enabled, our algorithm can improve the baseline method by a performance gain up to $11\%$.
The intuition can be illustrated by Fig. \ref{fig.ablationexample}.
The TF step can partially recover missing segments by utilizing information from neighboring frames, but the result
is coarse and false-positive outliers may be introduced in this step, as shown in Fig. \ref{fig.ablationexample.tf}.
Fortunately, the coarse segmentation can then be refined in the MR step to produce a high-quality result, as shown in Fig. \ref{fig.ablationexample.tfmr}.
Note that in this example, MR step itself cannot yield satisfactory results without the help of TF step for enlarging the positive labeling set,
as shown in Fig. \ref{fig.ablationexample.mr}.
To obtain the results in the rest of our experiments, we set the number of iterations $K=3$ for a good balance between  
performance and computational cost.


\subsection{Results}

\begin{table}[t!]\footnotesize
\begin{center}
 \begin{tabular}{cccccc}
\hline
\multirow{2}{*}{Method} & \hspace{6pt}Global\hspace{6pt} & \multicolumn{2}{c}{Region $\mathcal{J}$}    & \multicolumn{2}{c}{Contour $\mathcal{F}$} \\
    & Mean & Mean & Recall & Mean & Recall \\
\hline
\hline
Ours                                 & \textbf{0.675} & \textbf{0.645} & \textbf{0.741} & \textbf{0.705} & \underline{0.794}\\
apata\cite{DAVIS2017-2nd}            & \underline{0.666} & 0.634 & \underline{0.739} & \underline{0.699} & \textbf{0.801}\\
lixx\cite{DAVIS2017-1st}             & 0.661 & \underline{0.644} & 0.735 & 0.678 & 0.756\\
wangzhe		                         & 0.577 & 0.556 & 0.632 & 0.598 & 0.667\\
lalalaf--		                     & 0.574 & 0.545 & 0.613 & 0.602 & 0.688\\
voigtla--\cite{DAVIS2017-5th}        & 0.565 & 0.534 & 0.578 & 0.596 & 0.654\\
OnAVOS \cite{voigtlaender2017online} & 0.528 & 0.501 &  --   & 0.554 & --   \\
OSVOS \cite{caelles2017one}          & 0.505 & 0.472 & 0.508 & 0.537 & 0.578\\
\hline
\end{tabular}
\end{center}
\vspace{-5mm}
\caption{The results on the DAVIS 2017 test-dev set. The names ``apata'' and ``lixx'' are the top entries presented in the DAVIS 2017 Challenge.
The performances of OnAVOS and OSVOS, the top performers on DAVIS 2016, are shown for reference.
Note that both ``apata'' and ``lixx'' are heavy engineered systems that employ techniques
such as model ensembling, multi-scale training/testing, or even dedicated object (like person) detectors.}
\label{table.davis2017testdevres}
\vspace{-2mm}
\end{table}

\begin{table}[t!]\footnotesize
\begin{center}
 \begin{tabular}{cccc}
\hline
Method  & DAVIS 2016   &  Youtube-Objects  &  SegTrack v2\\
\hline
\hline
Ours                                    &  \underline{0.842}    &   \textbf{0.784}   & \underline{0.771} \\
OnAVOS \cite{voigtlaender2017online}    &  \textbf{0.857} &   0.774   &  -- \\
LucidTracker \cite{DAVIS2017-2nd}       &  0.837 &   0.762   & 0.768 \\
MaskRNN \cite{hu2017maskrnn}            &  0.804 &   --      & 0.721 \\
OSVOS \cite{caelles2017one}             &  0.798 &   \underline{0.783}   & 0.654 \\
MaskTrack \cite{perazzi2017learning}    &  0.797 &   0.726   & 0.703 \\
SegFlow \cite{cheng2017segflow}         &  0.761 &   --      & -- \\
STV \cite{wang2017super}                &  0.736 &   --      & \textbf{0.781} \\
CTN \cite{jang2017online}               &  0.735 &   --      & -- \\
VPN \cite{jampani2017cvpr}              &  0.702 &   --      & -- \\
PLM \cite{shin2017pixel}                &  0.700 &   --      & -- \\
ObjFlow \cite{tsai2016video}            &  0.680 &   0.776   & 0.741 \\
BVS \cite{marki2016bilateral}           &  0.600 &   --      & 0.584 \\
\hline
\end{tabular}
\end{center}
\vspace{-5mm}
\caption{The results on DAVIS 2016, Youtube-Objects and SegTrack v2 datasets. Only recent work in the semi-supervised setting (\ie, ground-truth of the first frame is given) is shown.
The results of DAVIS 2016 are evaluated on the validation set (scores mostly obtained from the DAVIS 2016 benchmark website \cite{davisbenchmarkwebsitemanual})
and only region IoU $\mathcal{J}$ score is shown. Our algorithm achieves state-of-the-art performances on all three datasets.}
\vspace{-4mm}
\label{table.traditionalbenchmarks}
\end{table}

\begin{figure*}[!t]
    \vspace{-3mm}
    \centerline{\subfloat{
        \includegraphics[width=0.16\textwidth]{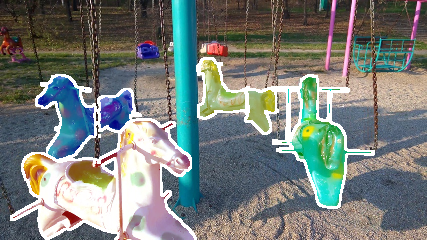}}
        \subfloat{
        \includegraphics[width=0.16\textwidth]{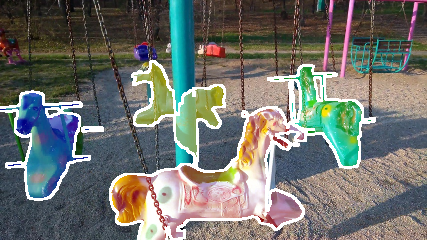}}
        \subfloat{
        \includegraphics[width=0.16\textwidth]{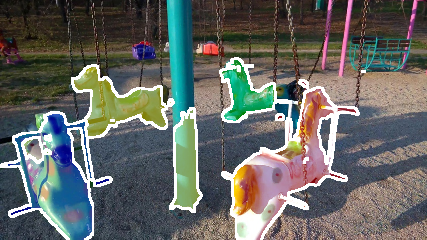}}
        \subfloat{
        \includegraphics[width=0.16\textwidth]{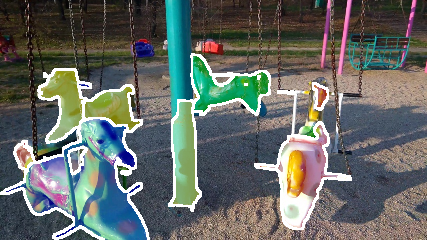}}
        \subfloat{
        \includegraphics[width=0.16\textwidth]{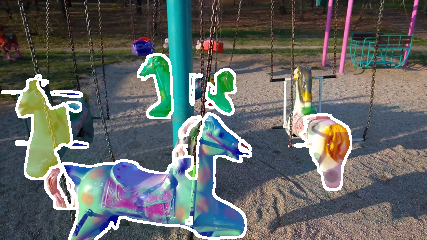}}
        \subfloat{
        \includegraphics[width=0.16\textwidth]{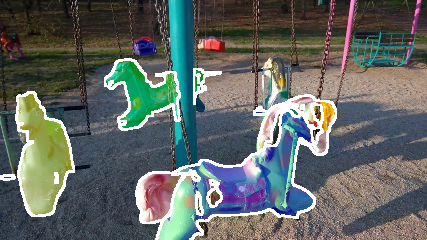}}}\vspace{-8pt}
    \centerline{\subfloat{
        \includegraphics[width=0.16\textwidth]{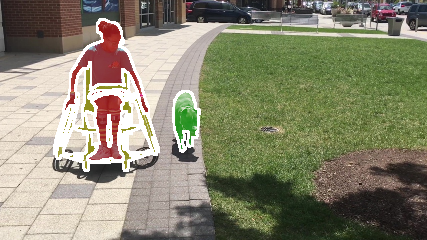}}
        \subfloat{
        \includegraphics[width=0.16\textwidth]{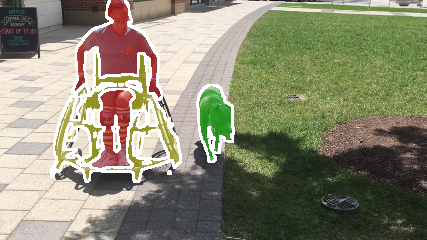}}
        \subfloat{
        \includegraphics[width=0.16\textwidth]{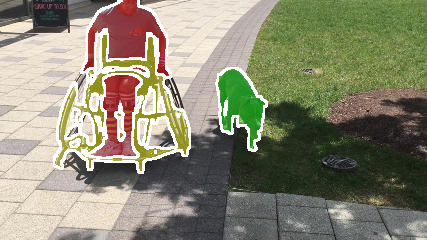}}
        \subfloat{
        \includegraphics[width=0.16\textwidth]{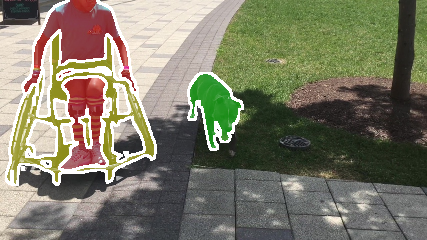}}
        \subfloat{
        \includegraphics[width=0.16\textwidth]{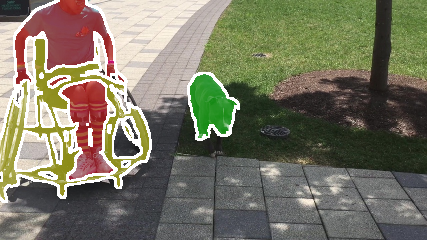}}
        \subfloat{
        \includegraphics[width=0.16\textwidth]{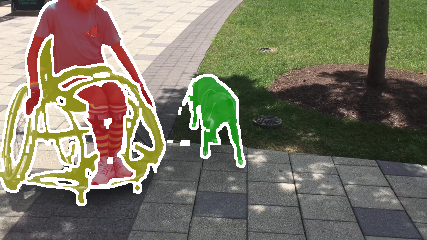}}}\vspace{-8pt}
    \centerline{\subfloat{
        \includegraphics[width=0.16\textwidth]{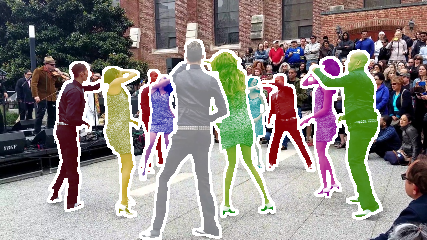}}
        \subfloat{
        \includegraphics[width=0.16\textwidth]{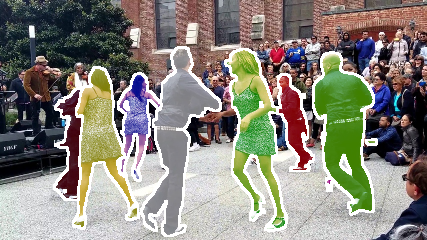}}
        \subfloat{
        \includegraphics[width=0.16\textwidth]{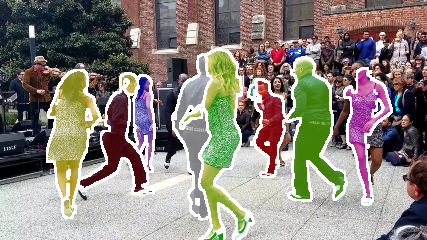}}
        \subfloat{
        \includegraphics[width=0.16\textwidth]{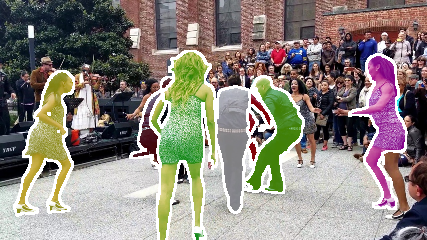}}
        \subfloat{
        \includegraphics[width=0.16\textwidth]{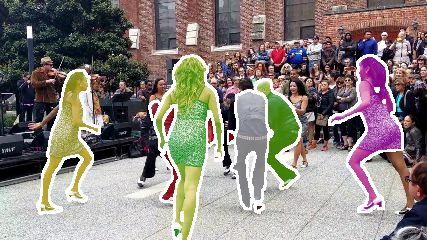}}
        \subfloat{
        \includegraphics[width=0.16\textwidth]{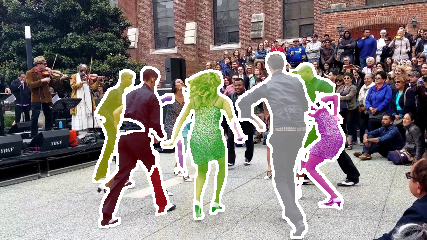}}}\vspace{-6pt}
    \caption{Examples of our results on the DAVIS 2017 test-dev set. The first column shows the ground-truth mask given in the first frame.
    The other columns are segmentation results for subsequent frames. From top to bottom, the sequences are ``carousel'', ``girl-dog'', and ``salsa'', respectively.
    Multiple objects are highlighted with different colors. It can be shown from these examples that the dataset is very challenging.
    Note that our algorithm does not employ specific object detectors
    (like a person detector used in a re-identification based method \cite{DAVIS2017-1st}) to achieve the results.}
    \vspace{-3.5mm}
    \label{fig.davis2017testdevvizres}
\end{figure*}

\begin{figure*}[!t]
    \vspace{-1mm}
    \centerline{\subfloat{
        \includegraphics[width=0.16\textwidth]{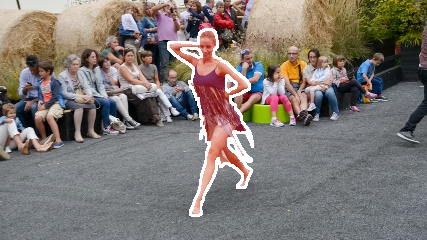}}
        \subfloat{
        \includegraphics[width=0.16\textwidth]{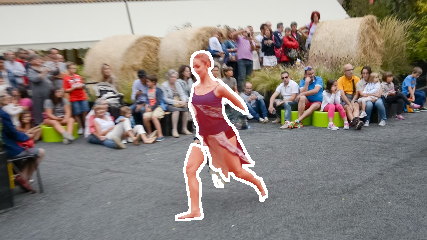}}
        \subfloat{
        \includegraphics[width=0.16\textwidth]{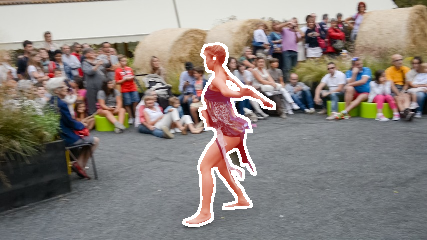}}
        \subfloat{
        \includegraphics[width=0.16\textwidth]{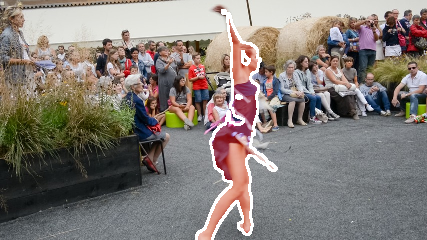}}
        \subfloat{
        \includegraphics[width=0.16\textwidth]{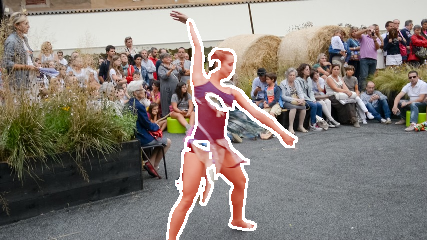}}
        \subfloat{
        \includegraphics[width=0.16\textwidth]{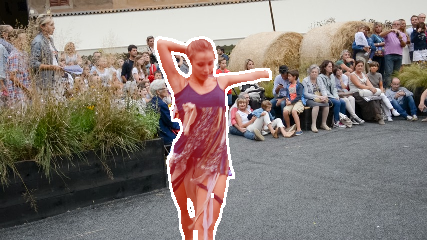}}}\vspace{-8pt}
    \centerline{\subfloat{
        \includegraphics[width=0.16\textwidth]{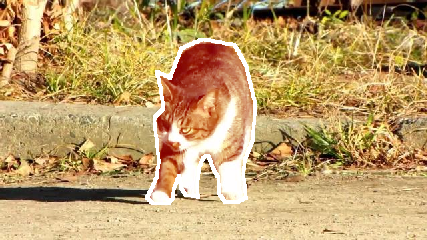}}
        \subfloat{
        \includegraphics[width=0.16\textwidth]{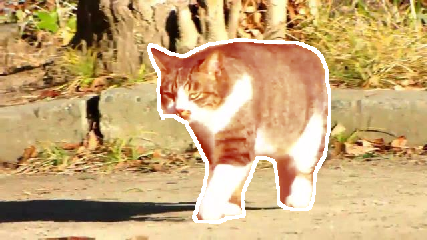}}
        \subfloat{
        \includegraphics[width=0.16\textwidth]{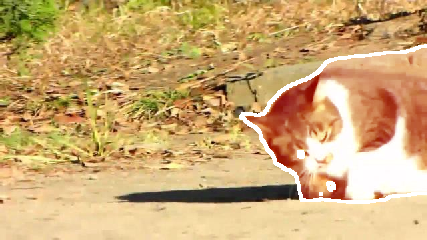}}
        \subfloat{
        \includegraphics[width=0.16\textwidth]{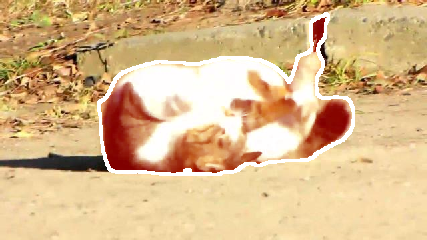}}
        \subfloat{
        \includegraphics[width=0.16\textwidth]{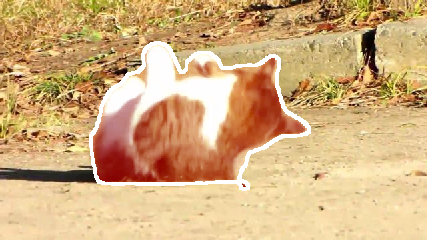}}
        \subfloat{
        \includegraphics[width=0.16\textwidth]{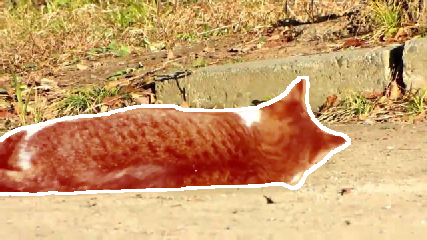}}}\vspace{-8pt}
    \centerline{\subfloat{
        \includegraphics[width=0.16\textwidth]{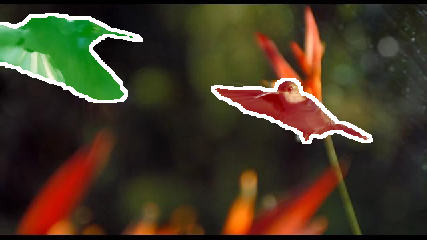}}
        \subfloat{
        \includegraphics[width=0.16\textwidth]{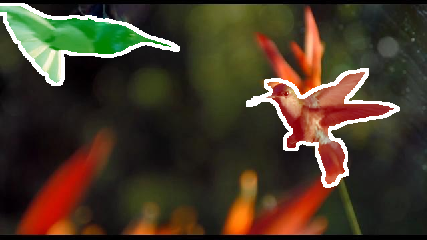}}
        \subfloat{
        \includegraphics[width=0.16\textwidth]{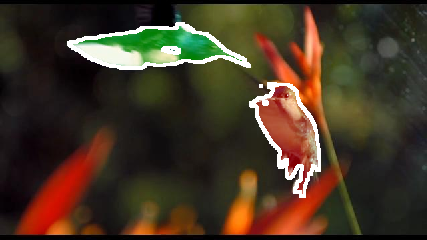}}
        \subfloat{
        \includegraphics[width=0.16\textwidth]{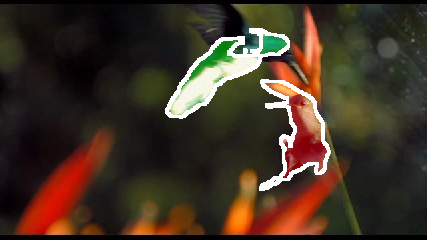}}
        \subfloat{
        \includegraphics[width=0.16\textwidth]{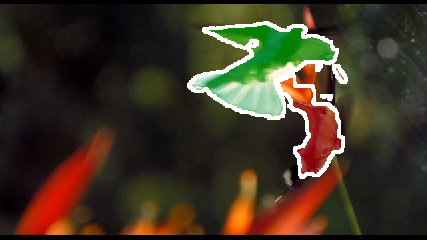}}
        \subfloat{
        \includegraphics[width=0.16\textwidth]{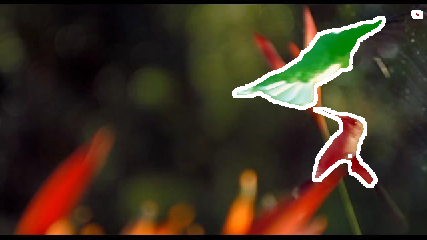}}}\vspace{-6pt}
    \caption{Examples of our results on DAVIS 2016, Youtube-Objects and SegTrack v2 datasets.
    From top to bottom, ``dance-twirl'' from DAVIS 2016, ``cat-0001'' from Youtube-Objects, and ``hummingbird'' from SegTrack v2, respectively.}
    \vspace{-3mm}
    \label{fig.tradbenchvizres}
\end{figure*}

We first report the performance of our algorithm on the challenging DAVIS 2017 test-dev set \cite{PontTuset2017davischallenge}.
The dataset consists of 30 video clips in various challenging cases including heavy occlusions, large appearance changes, complex shape deformation,
diverse object scales, \etc. It was used as a warm-up dataset in the DAVIS 2017 Challenge and remains open after the challenge.
Table \ref{table.davis2017testdevres} shows the performance of our algorithm comparing to the top-performing entries presented in the DAVIS 2017 Challenge.
Our algorithm outperforms the winning entries without resorting to techniques like model ensembling or multi-scale training/testing.
It also does not rely on dedicated object detectors, which makes it more general for class-agnostic object segmentation.
Fig. \ref{fig.davis2017testdevvizres} shows several examples of our segmentation results. Note that this dataset is very challenging and
methods purely relying on object appearance or methods mainly utilizing temporal information perform very poorly on the dataset.
For example, in the ``carousel'' sequence, appearance-based methods like OSVOS \cite{caelles2017one} will mistakenly switch object identities since the appearances
of all the carousels are very similar to each other when they are in a same orientation relative to camera.
On the other hand, methods like Object Flow \cite{tsai2016video} cannot handle occlusions due to lack of the ability to recognize re-appearing objects.
Our algorithm handles these cases well as shown in the first row of Fig. \ref{fig.davis2017testdevvizres},
thanks to the representation power of CNN for encoding object appearances and shapes
and the MRF modeling for establishing spatio-temporal connections.

For completeness, we also report the performance of our algorithm on three legacy datasets, DAVIS 2016 dataset \cite{perazzi2016benchmark},
Youtube-Objects dataset \cite{prest2012learning,jain2014supervoxel} and SegTrack v2 dataset \cite{li2013video},
as shown in Table \ref{table.traditionalbenchmarks} and Fig. \ref{fig.tradbenchvizres}.
With the same parameter settings as before, our algorithm achieves state-of-the-art results on all three datasets.
Note that these datasets are less challenging comparing to DAVIS 2017 and the scores tend to be saturated.
In Youtube-Objects and SegTrack v2, there are very few occlusions and appearance changes among the video sequences
and hence temporal propagation methods like Object Flow \cite{tsai2016video} can achieve very high performance.
In the more challenging DAVIS 2016 dataset, although there are large appearance changes and complex deformations in the sequences,
distractors and occlusions are much fewer than those in DAVIS 2017.
Most foreground objects can be correctly identified by CNN-based methods like
OSVOS \cite{caelles2017one} or its online adaptation version \cite{voigtlaender2017online},
without any temporal information leveraged.
By taking the advantages from both types of methods (temporal propagation and CNN),
our algorithm achieves state-of-the-art performance on all three datasets.

\vspace{-5pt}
\section{Conclusions}
\vspace{-5pt}

In this paper, we proposed a novel spatio-temporal MRF model for video object segmentation.
By performing inference in the MRF model, 
we developed an algorithm that alternates between a temporal fusion operation and a mask refinement feed-forward CNN, 
progressively inferring the results of video object segmentation.
We demonstrated the effectiveness of the proposed algorithm through extensive experiments on challenging datasets.
Different from previous efforts in combining MRFs and CNNs, our method explores the new direction
to embed a feed-forward pass of a CNN inside the inference of an MRF model. We hope that this idea could inspire more future work.

{\small
\bibliographystyle{ieee}
\bibliography{baoref_vos}
}

\end{document}